\documentclass[11pt,a4paper,usenames, dvipsnames]{article}
\usepackage{authblk}
\usepackage[hyperref]{naaclhlt2019}
\usepackage{times}
\usepackage{latexsym}
\usepackage{adjustbox}
\usepackage{booktabs}
\usepackage{dirtytalk}
\usepackage{longtable,tabu}
\usepackage{microtype}
\usepackage{url}

\aclfinalcopy % Uncomment this line for the final submission
\newcommand{\form}[1]{\say{#1}}
\newcommand{\lemma}[1]{\texttt{#1}}
\newcommand{\feats}[1]{\textsc{\MakeLowercase{#1}}}

\title{The SIGMORPHON 2019 Shared Task:\\ Morphological Analysis in Context and Cross-Lingual Transfer \\ for Inflection}

\newcommand{\s}[1]{\textsuperscript{#1}}

\DeclareSymbolFont{extraup}{U}{zavm}{m}{n}
\DeclareMathSymbol{\varheartsuit}{\mathalpha}{extraup}{86}
\DeclareMathSymbol{\vardiamondsuit}{\mathalpha}{extraup}{87}

\newcommand{\one}{\(\clubsuit\)}
\newcommand{\two}{\(\varheartsuit\)}
\newcommand{\three}{\(\spadesuit\)}
\newcommand{\four}{\(\vardiamondsuit\)}
\newcommand{\five}{\(\sharp\)}
\newcommand{\six}{\(\flat\)}
\newcommand{\seven}{\(\natural\)}
\newcommand{\annd}{\textnormal{, }}
\newcommand{\lastand}{\textnormal{, and }}

\author{Arya D. McCarthy\s{\one}%
  \annd Ekaterina Vylomova\s{\two}%
  \annd Shijie Wu\s{\one}%
  \annd 
       Chaitanya Malaviya\s{\three}%
  \annd \authorcr{} Lawrence Wolf-Sonkin\s{\four}%
  \annd Garrett Nicolai\s{\one}%
  \annd 
       Christo Kirov\s{\one}\thanks{~~Now at Google}%
  \annd Miikka Silfverberg\s{\five}%
  \annd \authorcr{} Sabrina J. Mielke\s{\one}%
  \annd
       Jeffrey Heinz\s{\six}%
  \annd Ryan Cotterell\s{\one}%
  \lastand Mans Hulden\s{\seven}\\
      \s{\one}Johns Hopkins University
\quad \s{\two}University of Melbourne
\quad \s{\three}Allen Institute for AI \\
      \s{\four}Google
\quad \s{\five}University of Helsinki
\quad \s{\six}Stony Brook University
\quad \s{\seven}University of Colorado
}

\date{}

\begin{document}
\maketitle

\thispagestyle{plain}
\pagestyle{plain}

\begin{abstract}
The SIGMORPHON 2019 shared task on cross-lingual transfer and contextual analysis in morphology examined transfer learning of inflection between 100 language pairs, as well as contextual lemmatization and morphosyntactic description in 66 languages. The first task evolves past years' inflection tasks by examining transfer of morphological inflection knowledge from a high-resource language to a low-resource language. This year also presents a new second challenge on lemmatization and morphological feature analysis in context. All submissions featured a neural component and built on either this year's strong baselines or highly ranked systems from previous years' shared tasks. Every participating team improved in accuracy over the baselines for the inflection task (though not Levenshtein distance), and every team in the contextual analysis task improved on both state-of-the-art neural and non-neural baselines.
\end{abstract}

\section{Introduction}
While producing a sentence, humans 
combine various types of knowledge
to produce fluent output---various shades of
meaning are expressed through word selection and
tone, while the language is made to conform to
underlying structural rules via syntax and 
morphology.  Native speakers are often quick
to identify disfluency, even if the meaning
of a sentence is mostly clear.

Automatic systems must also consider these 
constraints when constructing or processing language.
Strong enough language models can often reconstruct
common syntactic structures, but are
insufficient to properly model morphology.
Many languages implement large inflectional 
paradigms that mark both function and content words
with a varying levels of morphosyntactic information. For instance, Romanian verb forms inflect for person, number, tense, mood, and voice; meanwhile, Archi verbs can take on thousands of forms \citep{kibrik1998archi}. Such complex paradigms 
produce large inventories of words, all of which
must be producible by a realistic system, even
though a large percentage of them will never
be observed over billions of lines of linguistic
input.  
Compounding the issue, good inflectional systems
often require large amounts of supervised training
data, which is infeasible in many of the world's
languages.

This year's shared task is concentrated on
encouraging the construction of strong morphological
systems that perform two related but
different inflectional
tasks.  The first task asks
participants to create
morphological inflectors for a large number of 
under-resourced languages, encouraging systems
that use highly-resourced, related languages as
a cross-lingual training signal.  
The second task 
welcomes submissions that
invert this operation in light of contextual information: Given an
unannotated sentence, lemmatize each word, and
tag them with a morphosyntactic description.
Both of these tasks extend upon previous
morphological competitions, and the best submitted
systems now represent the state of the art in 
their respective tasks.

\section{Tasks and Evaluation}

\subsection{Task 1: Cross-lingual transfer for morphological inflection}

Annotated resources for the world's languages are not distributed equally---some languages simply have more as they have more native speakers willing and able to annotate more data. We explore how to transfer knowledge from high-resource languages that are genetically related to low-resource languages.

The first task iterates on last year's main task: morphological inflection \citep{cotterell-etal-2018-conll}. Instead of giving some number of training examples in the language of interest, we provided only a limited number in that language. To accompany it, we provided a larger number of examples in either a related or unrelated language.
Each test example asked participants to produce some other inflected form when given a lemma and a bundle of morphosyntactic features as input.
The goal, thus, is to perform morphological inflection in the low-resource language, having hopefully exploited some similarity to the high-resource language.
Models which perform well here can aid downstream tasks like machine translation in low-resource settings. All datasets were resampled from UniMorph, which makes them distinct from past years.

The mode of the task is inspired by \citet{zoph-etal-2016-transfer}, who fine-tune a model pre-trained on a high-resource language to perform well on a low-resource language. We do not, though, require that models be trained by fine-tuning. Joint modeling or any number of methods may be explored instead.

\paragraph{Example}
The model will have access to type-level data in a low-resource target language, plus a high-resource source language. We give an example here of Asturian as the target language with Spanish as the source language.

\bigskip
\begin{table}[h]
\begin{adjustbox}{width=\linewidth}
\begin{tabular}{l l l}
\multicolumn{3}{p{\linewidth}}{\textbf{Low-resource target training data (Asturian)}} \\
\lemma{facer} &     \form{fechu}    &  \feats{V;V.PTCP;PST} \\
\lemma{aguar} &     \form{agu\`{a}} &      \feats{V;PRS;2;PL;IND} \\
\vdots & \vdots & \vdots \\
\multicolumn{3}{p{\linewidth}}{\textbf{High-resource source language training data (Spanish)}} \\
\lemma{tocar}   & \form{tocando}    & \feats{V;V.PTCP;PRS} \\
\lemma{bailar}  & \form{bailaba}    & \feats{V;PST;IPFV;3;SG;IND} \\ 
\lemma{mentir}  & \form{minti\'{o}} &  \feats{V;PST;PFV;3;SG;IND} \\
\vdots & \vdots & \vdots \\
\multicolumn{3}{p{\linewidth}}{\textbf{Test input (Asturian)}} \\
\lemma{baxar}   & & \feats{V;V.PTCP;PRS} \\
\multicolumn{3}{p{\linewidth}}{\textbf{Test output (Asturian)}} \\
& \form{baxando}\\
\end{tabular}
\end{adjustbox}
\caption{Sample language pair and data format for Task~1}
\label{tab:sub1data}
\end{table}

\paragraph{Evaluation} We score the output of each system in terms of its predictions' exact-match accuracy and the average Levenshtein distance between the predictions and their corresponding true forms.

\subsection{Task 2: Morphological analysis in context}

Although inflection of words in a context-agnostic
manner is a useful evaluation of the morphological
quality of a system, people do not learn morphology in isolation.

In 2018, the second task of the 
CoNLL--SIGMORPHON Shared
Task~\cite{cotterell-etal-2018-conll} required submitting 
systems to complete an inflectional cloze task \citep{taylor1953cloze} given
only the sentential context and the desired lemma -- an example of the
problem is given in the following lines:
A successful system would predict the plural form ``dogs''. Likewise, a Spanish word form \form{ayuda} may be a feminine noun or a third-person verb form, which must be disambiguated by context.

\bigskip

\begin{tabular}{l l l }
The & \underline{\phantom{(dog)}} & are barking. \\
 & (dog) &
\end{tabular}{}

\bigskip

This year's task extends the second task from last
year.  Rather than inflect a single word in context,
the task is to provide a complete morphological
tagging of a sentence: for each word, a successful
system will need to lemmatize and tag it with
a morphsyntactic description (MSD).

\bigskip

\begin{adjustbox}{width=\linewidth}
\begin{tabular}{l l l l l}
\toprule
The & dogs & are & barking &. \\
\midrule
the & dog & be & bark & . \\
DET & N;PL & V;PRS;3;PL & V;V.PTCP;PRS & PUNCT \\
\bottomrule
\end{tabular}{}
\end{adjustbox}

\bigskip

Context is critical---depending on the sentence,
identical word forms realize a large
number of potential inflectional categories, which
will in turn influence lemmatization decisions.
If the sentence were instead ``The barking dogs
kept us up all night'', ``barking'' is now an
adjective, and its lemma is also ``barking''.

\section{Data}

\subsection{Data for Task 1}

\paragraph{Language pairs}
We presented data in 100 language pairs spanning 79 unique languages. Data for all but four languages (Basque, Kurmanji, Murrinhpatha, and Sorani) are extracted from English Wiktionary, a large multi-lingual crowd-sourced dictionary with morphological paradigms for many lemmata.\footnote{The Basque language data was extracted from a manually designed finite-state morphological analyzer \citep{alegria2009porting}. Murrinhpatha data was donated by John Mansfield; it is discussed in \citet{mansfield2019murrinhpatha}. Data for Kurmanji Kurdish and Sorani Kurdish were created as part of the Alexina project \citep{walther2010fast, walther2010developing}.} 20 of the 100 language pairs are either distantly related or unrelated; this allows speculation into the relative importance of data quantity and linguistic relatedness.

\paragraph{Data format}
For each language, the basic data consists of triples of the form
(lemma, feature bundle, inflected form), as in \autoref{tab:sub1data}.
The first feature in the bundle always specifies the core part of
speech (e.g., verb). For each language pair, separate files contain the high- and low-resource training examples.

  All features in the
bundle are coded according to the UniMorph Schema, a
cross-linguistically consistent universal morphological feature set
\cite{sylak-glassmankirov2015,sylakglassman-EtAl:2015:ACL-IJCNLP}.

\paragraph{Extraction from Wiktionary}

For each of the Wiktionary languages, Wiktionary provides a number
of tables, each of which specifies the full inflectional paradigm for
a particular lemma.  As in the previous iteration, tables were extracted using a template annotation procedure described in \citep{kirov-etal-2018-unimorph}.

\paragraph{Sampling data splits}

From each language's collection of paradigms, we sampled the training, development, and test sets as in 2018.\footnote{These datasets can be obtained from \url{https://sigmorphon.github.io/sharedtasks/2019/}} Crucially, while the data were sampled in the same fashion, the datasets are distinct from those used for the 2018 shared task.

Our first step was to construct probability distributions over the (lemma, feature bundle, inflected form) triples in our full dataset.
For each triple, we counted how many tokens the inflected form has in the February 2017 dump of Wikipedia for that language.
To distribute the counts of an observed form over all the triples that have this token as its form, we follow the method used in the previous shared task \citep{cotterell-etal-2018-conll}, training a neural network on unambiguous forms to estimate the distribution over all, even ambiguous, forms.
We then sampled 12,000 triples without replacement from this distribution.  The first 100 were taken as training data for low-resource settings. The first 10,000 were used as high-resource training sets. As these sets are nested, the highest-count triples tend to appear in the smaller training sets.\footnote{Several high-resource languages had necessarily fewer, but on a similar order of magnitude. Bengali, Uzbek, Kannada, Swahili. Likewise, the low-resource language Telugu had fewer than 100 forms.}

The final 2000 triples were randomly shuffled and then split in half to obtain development and test sets of 1000 forms each.\footnote{When sufficient data are unavailable, we instead use 50 or 100 examples.} The final shuffling was performed to ensure that the development set is similar to the test set.  By contrast, the  development and test sets tend to contain lower-count triples than the training set.\footnote{This mimics a realistic setting, as supervised training is usually employed to generalize from frequent words that appear in annotated resources to less frequent words that do not.  Unsupervised learning methods also tend to generalize from more frequent words (which can be analyzed more easily by combining information from many contexts) to less frequent ones.}
% In those languages where we have less than 12000 total forms, we omit the high-resource training set (all languages have at least 3000 forms).
%Note that for languages that do not have enough triples for this process, we settle for omitting the higher-resource training regimes and scale down the other sizes. Details for all languages are found in \autoref{tab:dqp1,tab:dqp2}.

\paragraph{Other modifications}
We further adopted some changes to increase compatibility. Namely, we corrected some annotation errors created while scraping Wiktionary for the 2018 task, and we standardized Romanian t-cedilla and t-comma to t-comma. (The same was done with s-cedilla and s-comma.)

\subsection{Data for Task 2}
Our data for task 2 come from the Universal Dependencies treebanks \citep[UD;][v2.3]{nivre2018universal}, which provides pre-defined training, development, and test splits and annotations in a unified annotation schema for morphosyntax and dependency relationships. Unlike the 2018 cloze task which used UD data, we require no manual data preparation and are able to leverage all 107 monolingual treebanks. As is typical, data are presented in CoNLL-U format,\footnote{\url{https://universaldependencies.org/format.html}} although we modify the morphological feature and lemma fields.

\paragraph{Data conversion}
The morphological annotations for the 2019 shared task were converted to the UniMorph schema \citep{kirov-etal-2018-unimorph} according to 
\citet{mccarthy-etal-2018-marrying}, who provide a deterministic mapping that increases agreement across languages. This also moves the part of speech into the bundle of morphological features. We do not attempt to individually correct any errors in the UD source material. Further, some languages received additional pre-processing. In the Finnish data, we removed morpheme boundaries that were present in the lemmata (e.g., \lemma{puhe\#kieli} \(\mapsto\) \lemma{puhekieli} `spoken+language'). Russian lemmata in the GSD treebank were presented in all uppercase; to match the 2018 shared task, we lowercased these. In development and test data, all fields except for form and index within the sentence were struck.

\section{Baselines}

\subsection{Task 1 Baseline}

We include four neural sequence-to-sequence models mapping lemma into inflected word forms: soft attention \cite{luong2015effective}, non-monotonic hard attention \cite{wu2018hard}, monotonic hard attention and a variant with offset-based transition distribution \cite{wu2019exact}.
Neural sequence-to-sequence models with soft attention \cite{luong2015effective} have dominated previous SIGMORPHON shared tasks \cite{cotterell2017conll}.
\citet{wu2018hard} instead models the alignment between characters in the lemma and the inflected word form explicitly with hard attention and learns this alignment and transduction jointly.
\citet{wu2019exact} shows that enforcing strict monotonicity with hard attention is beneficial in tasks such as morphological inflection where the transduction is mostly monotonic.
The encoder is a biLSTM while the decoder is a left-to-right LSTM. All models use multiplicative attention and have roughly the same number of parameters. In the model, a morphological tag is fed to the decoder along with target character embeddings to guide the decoding.  During the training of the hard attention model, dynamic programming is applied to marginalize all latent alignments exactly.
% \citep{wu2019exact}: 

\subsection{Task 2 Baselines}

\paragraph{Non-neural} \citep{muller-etal-2015-joint}: The Lemming model is a log-linear model that performs joint morphological tagging and lemmatization. The model is globally normalized with the use of a second order linear-chain CRF. To efficiently calculate the partition function, the choice of lemmata are pruned with the use of pre-extracted edit trees.

\paragraph{Neural} \citep{malaviya-wu-2019simple}: This is a state-of-the-art neural model that also performs joint morphological tagging and lemmatization, but also accounts for the exposure bias with the application of maximum likelihood (MLE). The model stitches the tagger and lemmatizer together with the use of jackknifing \cite{agic-schluter-2017-train} to expose the lemmatizer to the errors made by the tagger model during training. The morphological tagger is based on a character-level biLSTM embedder that produces the embedding for a word, and a word-level biLSTM tagger that predicts a morphological tag sequence for each word in the sentence. The lemmatizer is a neural sequence-to-sequence model \cite{wu2019exact} that uses the decoded morphological tag sequence from the tagger as an additional attribute. The model uses hard monotonic attention instead of standard soft attention, along with a dynamic programming based training scheme.

\section{Results}

The SIGMORPHON 2019 shared task received 30 submissions---14 for task 1 and 16 for task 2---from 23 teams. 
In addition, the organizers' baseline systems were evaluated. % Can delete for space concerns.

\subsection{Task 1 Results}

\begin{table}
    \centering
    \begin{adjustbox}{max width=\linewidth}
    \begin{tabular}{l r r}
\toprule
Team & Avg.\ Accuracy & Avg.\ Levenshtein \\
\midrule

AX-01 & 18.54 & 3.62 \\
AX-02 & 24.99 & 2.72 \\
CMU-03 & \textbf{58.79} & 1.52 \\
IT-IST-01 & 49.00 & \textbf{1.29} \\
IT-IST-02 & 50.18 & 1.32 \\
Tuebingen-01$\dagger$ & 34.49 & 1.88 \\
Tuebingen-02$\dagger$ & 20.86 & 2.36 \\
UAlberta-01* & 48.33 & 1.23 \\
UAlberta-02*$\dagger$ & 54.75 & 1.03 \\
UAlberta-03*$\dagger$ & 8.45 & 4.06 \\
UAlberta-04*$\dagger$ & 11.00 & 3.86 \\
UAlberta-05* & 4.10 & 3.08 \\
UAlberta-06*$\dagger$ & 26.85 & 2.65 \\ \midrule
Baseline & 48.55 & 1.33 \\
\bottomrule

    \end{tabular}
    \end{adjustbox}
    \caption{Task 1 Team Scores, averaged across all Languages; * indicates submissions were
    only applied to a subset of languages, making scores incomparable. $\dagger$ indicates that additional resources were used for training.}
\end{table}

\begin{table*}
    \centering
    \begin{adjustbox}{max width=\textwidth}
    \begin{tabular}{l r r l | l r r l}
\toprule
HRL--LRL & Baseline & Best & Team & HRL--LRL & Baseline & Best & Team \\
\midrule
adyghe--kabardian & 96.0 & 97.0 & Tuebingen-02 &       hungarian--livonian & 29.0 & 44.0 & it-ist-01 \\
albanian--breton & 40.0 & 81.0 & CMU-03 &       hungarian--votic & 19.0 & 34.0 & it-ist-01 \\
arabic--classical-syriac & 66.0 & 92.0 & CMU-03 &       irish--breton & 39.0 & 79.0 & CMU-03 \\
arabic--maltese & 31.0 & 41.0 & CMU-03 &       irish--cornish & 24.0 & 34.0 & it-ist-01 \\
arabic--turkmen & 74.0 & 84.0 & CMU-03 &       irish--old-irish & 2.0 & 6.0 & it-ist-02 \\
armenian--kabardian & 83.0 & 87.0 & it-ist-01 &       irish--scottish-gaelic & 64.0 & 66.0 & CMU-03 \\
asturian--occitan & 48.0 & 77.0 & CMU-03 &       italian--friulian & 56.0 & 78.0 & CMU-03 \\
bashkir--azeri & 39.0 & 69.0 & it-ist-02 &       italian--ladin & 55.0 & 74.0 & CMU-03 \\
bashkir--crimean-tatar & 70.0 & 70.0 & CMU-03 &       italian--maltese & 26.0 & 45.0 & CMU-03 \\
bashkir--kazakh & 80.0 & 90.0 & it-ist-01 &       italian--neapolitan & 80.0 & 83.0 & CMU-03 \\
bashkir--khakas & 86.0 & 96.0 & it-ist-02 &       kannada--telugu & 82.0 & 94.0 & CMU-03 \\
bashkir--tatar & 68.0 & 74.0 & it-ist-02 &       kurmanji--sorani & 15.0 & 69.0 & CMU-03 \\
bashkir--turkmen & 94.0 & 88.0 & it-ist-01 &       latin--czech & 20.1 & 71.4 & CMU-03 \\
basque--kashubian & 40.0 & 76.0 & CMU-03 &       latvian--lithuanian & 17.1 & 48.4 & CMU-03 \\
belarusian--old-irish & 2.0 & 10.0 & CMU-03 &       latvian--scottish-gaelic & 48.0 & 68.0 & CMU-03 \\
bengali--greek & 17.7 & 74.6 & CMU-03 &       persian--azeri & 46.0 & 69.0 & CMU-03 \\
bulgarian--old-church-slavonic & 44.0 & 56.0 & CMU-03 &       persian--pashto & 27.0 & 48.0 & CMU-03 \\
czech--kashubian & 52.0 & 78.0 & CMU-03 &       polish--kashubian & 74.0 & 78.0 & CMU-03 \\
czech--latin & 8.4 & 42.0 & CMU-03 &       polish--old-church-slavonic & 40.0 & 58.0 & CMU-03 \\
danish--middle-high-german & 72.0 & 82.0 & it-ist-02 &       portuguese--russian & 27.5 & 76.3 & CMU-03 \\
danish--middle-low-german & 36.0 & 44.0 & it-ist-01 &       romanian--latin & 6.7 & 41.3 & CMU-03 \\
danish--north-frisian & 28.0 & 46.0 & CMU-03 &       russian--old-church-slavonic & 34.0 & 64.0 & CMU-03 \\
danish--west-frisian & 42.0 & 43.0 & CMU-03 &       russian--portuguese & 50.5 & 88.4 & CMU-03 \\
danish--yiddish & 76.0 & 67.0 & it-ist-01 &       sanskrit--bengali & 33.0 & 65.0 & CMU-03 \\
dutch--middle-high-german & 76.0 & 78.0 & it-ist-01 / it-ist-02 &       sanskrit--pashto & 34.0 & 43.0 & CMU-03 \\
dutch--middle-low-german & 42.0 & 52.0 & it-ist-02 &       slovak--kashubian & 54.0 & 76.0 & CMU-03 \\
dutch--north-frisian & 32.0 & 46.0 & CMU-03 &       slovene--old-saxon & 10.6 & 53.2 & CMU-03 \\
dutch--west-frisian & 38.0 & 51.0 & it-ist-02 &       sorani--irish & 27.6 & 66.3 & CMU-03 \\
dutch--yiddish & 78.0 & 64.0 & it-ist-01 &       spanish--friulian & 53.0 & 81.0 & CMU-03 \\
english--murrinhpatha & 22.0 & 42.0 & it-ist-02 &       spanish--occitan & 57.0 & 78.0 & CMU-03 \\
english--north-frisian & 31.0 & 42.0 & CMU-03 &       swahili--quechua & 13.9 & 92.1 & CMU-03 \\
english--west-frisian & 35.0 & 43.0 & CMU-03 &       turkish--azeri & 80.0 & 87.0 & it-ist-02 \\
estonian--ingrian & 30.0 & 44.0 & it-ist-02 &       turkish--crimean-tatar & 83.0 & 89.0 & CMU-03 / it-ist-02 \\
estonian--karelian & 74.0 & 68.0 & it-ist-01 &       turkish--kazakh & 76.0 & 86.0 & it-ist-02 \\
estonian--livonian & 36.0 & 40.0 & it-ist-02 &       turkish--khakas & 76.0 & 94.0 & it-ist-01 \\
estonian--votic & 25.0 & 35.0 & it-ist-01 &       turkish--tatar & 73.0 & 83.0 & it-ist-02 \\
finnish--ingrian & 54.0 & 48.0 & it-ist-02 &       turkish--turkmen & 86.0 & 98.0 & it-ist-01 \\
finnish--karelian & 70.0 & 78.0 & it-ist-01 &       urdu--bengali & 49.0 & 67.0 & CMU-03 \\
finnish--livonian & 22.0 & 34.0 & CMU-03 / it-ist-01 &       urdu--old-english & 20.8 & 40.3 & CMU-03 \\
finnish--votic & 42.0 & 40.0 & it-ist-02 &       uzbek--azeri & 57.0 & 70.0 & CMU-03 \\
french--occitan & 50.0 & 80.0 & CMU-03 &       uzbek--crimean-tatar & 67.0 & 67.0 & CMU-03 \\
german--middle-high-german & 72.0 & 82.0 & CMU-03 &       uzbek--kazakh & 84.0 & 72.0 & CMU-03 \\
german--middle-low-german & 42.0 & 52.0 & it-ist-02 &       uzbek--khakas & 86.0 & 92.0 & it-ist-01 \\
german--yiddish & 77.0 & 68.0 & it-ist-01 &       uzbek--tatar & 69.0 & 72.0 & CMU-03 \\
greek--bengali & 51.0 & 67.0 & CMU-03 &       uzbek--turkmen & 80.0 & 78.0 & CMU-03 \\
hebrew--classical-syriac & 89.0 & 95.0 & CMU-03 &       welsh--breton & 45.0 & 86.0 & CMU-03 \\
hebrew--maltese & 37.0 & 47.0 & CMU-03 &       welsh--cornish & 22.0 & 42.0 & it-ist-01 \\
hindi--bengali & 54.0 & 68.0 & CMU-03 &       welsh--old-irish & 6.0 & 6.0 & CMU-03 \\
hungarian--ingrian & 12.0 & 40.0 & it-ist-01 &       welsh--scottish-gaelic & 40.0 & 64.0 & CMU-03 \\
hungarian--karelian & 62.0 & 70.0 & it-ist-02 &       zulu--swahili & 44.0 & 81.0 & CMU-03 \\\bottomrule

    \end{tabular}
    \end{adjustbox}
    \caption{Task 1 Accuracy scores}
\end{table*}

\begin{table*}
    \centering
    \begin{adjustbox}{max width=\textwidth}
    \begin{tabular}{l r r l | l r r l}
\toprule
HRL--LRL & Baseline & Best & Team & HRL--LRL & Baseline & Best & Team \\
\midrule

adyghe--kabardian & 0.04 & 0.03 & Tuebingen-02 &                    hungarian--livonian & 2.56 & 1.81 & it-ist-02 \\
albanian--breton & 1.30 & 0.44 & it-ist-02 &                    hungarian--votic & 2.47 & 1.11 & it-ist-01 \\
arabic--classical-syriac & 0.46 & 0.10 & CMU-03 &                    irish--breton & 1.57 & 0.38 & CMU-03 \\
arabic--maltese & 1.42 & 1.37 & CMU-03 &                    irish--cornish & 2.00 & 1.56 & it-ist-01 \\
arabic--turkmen & 0.46 & 0.32 & CMU-03 &                    irish--old-irish & 3.30 & 3.12 & it-ist-02 \\
armenian--kabardian & 0.21 & 0.14 & CMU-03 / it-ist-01 &                    irish--scottish-gaelic & 0.96 & 1.06 & CMU-03 \\
asturian--occitan & 1.74 & 0.80 & it-ist-01 &                    italian--friulian & 1.03 & 0.72 & it-ist-02 \\
bashkir--azeri & 1.64 & 0.69 & it-ist-02 &                    italian--ladin & 0.79 & 0.60 & CMU-03 \\
bashkir--crimean-tatar & 0.39 & 0.42 & CMU-03 &                    italian--maltese & 1.39 & 1.23 & CMU-03 \\
bashkir--kazakh & 0.32 & 0.10 & it-ist-01 &                    italian--neapolitan & 0.40 & 0.36 & it-ist-02 \\
bashkir--khakas & 0.18 & 0.04 & it-ist-02 &                    kannada--telugu & 0.60 & 0.14 & CMU-03 \\
bashkir--tatar & 0.46 & 0.33 & CMU-03 &                    kurmanji--sorani & 2.56 & 0.65 & CMU-03 \\
bashkir--turkmen & 0.10 & 0.12 & it-ist-01 &                    latin--czech & 2.77 & 1.14 & CMU-03 \\
basque--kashubian & 1.16 & 0.42 & CMU-03 &                    latvian--lithuanian & 2.21 & 1.69 & CMU-03 \\
belarusian--old-irish & 3.90 & 3.14 & CMU-03 &                    latvian--scottish-gaelic & 1.16 & 1.00 & CMU-03 \\
bengali--greek & 2.86 & 0.59 & CMU-03 &                    persian--azeri & 1.35 & 0.74 & CMU-03 \\
bulgarian--old-church-slavonic & 1.14 & 1.06 & CMU-03 &                    persian--pashto & 1.70 & 1.54 & CMU-03 \\
czech--kashubian & 0.84 & 0.36 & CMU-03 &                    polish--kashubian & 0.34 & 0.34 & CMU-03 \\
czech--latin & 2.95 & 1.36 & CMU-03 &                    polish--old-church-slavonic & 1.22 & 0.96 & CMU-03 \\
danish--middle-high-german & 0.50 & 0.38 & it-ist-02 &                    portuguese--russian & 1.70 & 1.16 & CMU-03 \\
danish--middle-low-german & 1.44 & 1.26 & it-ist-01 &                    romanian--latin & 3.05 & 1.35 & CMU-03 \\
danish--north-frisian & 2.78 & 2.11 & CMU-03 &                    russian--old-church-slavonic & 1.33 & 0.86 & CMU-03 \\
danish--west-frisian & 1.57 & 1.27 & it-ist-02 &                    russian--portuguese & 1.04 & 0.66 & CMU-03 \\
danish--yiddish & 0.91 & 0.72 & Tuebingen-01 &                    sanskrit--bengali & 1.79 & 1.13 & CMU-03 \\
dutch--middle-high-german & 0.44 & 0.36 & it-ist-02 &                    sanskrit--pashto & 1.54 & 1.27 & it-ist-02 \\
dutch--middle-low-german & 1.34 & 1.16 & it-ist-02 &                    slovak--kashubian & 0.60 & 0.34 & CMU-03 \\
dutch--north-frisian & 2.67 & 1.99 & CMU-03 &                    slovene--old-saxon & 2.23 & 1.14 & CMU-03 \\
dutch--west-frisian & 2.18 & 1.18 & it-ist-02 &                    sorani--irish & 2.40 & 0.99 & CMU-03 \\
dutch--yiddish & 0.53 & 0.72 & Tuebingen-01 &                    spanish--friulian & 1.01 & 0.61 & CMU-03 \\
english--murrinhpatha & 1.68 & 1.10 & it-ist-02 &                    spanish--occitan & 1.14 & 0.57 & it-ist-01 \\
english--north-frisian & 2.73 & 2.22 & it-ist-02 &                    swahili--quechua & 3.90 & 0.56 & CMU-03 \\
english--west-frisian & 1.48 & 1.26 & it-ist-02 &                    turkish--azeri & 0.35 & 0.22 & it-ist-01 \\
estonian--ingrian & 1.56 & 1.24 & it-ist-02 &                    turkish--crimean-tatar & 0.24 & 0.14 & CMU-03 \\
estonian--karelian & 0.52 & 0.62 & it-ist-02 &                    turkish--kazakh & 0.34 & 0.16 & it-ist-02 \\
estonian--livonian & 1.87 & 1.47 & it-ist-02 &                    turkish--khakas & 0.80 & 0.06 & it-ist-01 \\
estonian--votic & 1.55 & 1.17 & it-ist-02 &                    turkish--tatar & 0.37 & 0.21 & it-ist-02 \\
finnish--ingrian & 1.08 & 1.20 & it-ist-02 &                    turkish--turkmen & 0.24 & 0.02 & it-ist-01 \\
finnish--karelian & 0.64 & 0.42 & it-ist-01 &                    urdu--bengali & 1.12 & 0.98 & CMU-03 \\
finnish--livonian & 2.48 & 1.71 & it-ist-01 &                    urdu--old-english & 1.72 & 1.20 & CMU-03 \\
finnish--votic & 1.25 & 1.02 & it-ist-02 &                    uzbek--azeri & 1.23 & 0.70 & CMU-03 \\
french--occitan & 1.22 & 0.69 & it-ist-01 &                    uzbek--crimean-tatar & 0.49 & 0.45 & CMU-03 \\
german--middle-high-german & 0.44 & 0.32 & it-ist-02 &                    uzbek--kazakh & 0.20 & 0.32 & CMU-03 \\
german--middle-low-german & 1.24 & 1.16 & it-ist-02 &                    uzbek--khakas & 0.24 & 0.18 & it-ist-01 \\
german--yiddish & 0.46 & 0.72 & Tuebingen-01 &                    uzbek--tatar & 0.48 & 0.35 & CMU-03 \\
greek--bengali & 1.21 & 1.02 & CMU-03 &                    uzbek--turkmen & 0.32 & 0.42 & CMU-03 \\
hebrew--classical-syriac & 0.14 & 0.06 & CMU-03 &                    welsh--breton & 0.90 & 0.31 & CMU-03 \\
hebrew--maltese & 1.24 & 1.10 & CMU-03 &                    welsh--cornish & 2.44 & 1.50 & it-ist-01 \\
hindi--bengali & 1.18 & 0.72 & UAlberta-02 &                    welsh--old-irish & 3.36 & 3.08 & CMU-03 \\
hungarian--ingrian & 2.60 & 1.46 & it-ist-01 &                    welsh--scottish-gaelic & 1.22 & 1.08 & CMU-03 \\
hungarian--karelian & 0.90 & 0.50 & it-ist-01 &                    zulu--swahili & 1.24 & 0.33 & CMU-03 \\\bottomrule

    \end{tabular}
    \end{adjustbox}
    \caption{Task 1 Levenshtein scores}
\end{table*}

Five teams participated in the first Task, with 
a variety of methods aimed at leveraging the 
cross-lingual data to improve system performance.

The University of Alberta (UAlberta) performed a focused 
investigation on four language pairs, training
cognate-projection systems from external cognate
lists.  Two methods were considered: 
one which trained a 
high-resource neural encoder-decoder, and projected
the test data into the HRL, and one that projected
the HRL data into the LRL, and trained a combined
system.  Results demonstrated that certain 
language pairs may be amenable to such methods.

The Tuebingen University submission (Tuebingen) aligned source and 
target to learn a set of edit-actions with both
linear and neural classifiers that independently
learned to predict action sequences for each
morphological category.  Adding in the cross-lingual
data only led to modest gains.

AX-Semantics combined the low- and
high-resource data to train an encoder-decoder 
seq2seq model; optionally also implementing
domain adaptation methods to focus later epochs
on the target language.

The CMU submission first attends over 
a decoupled representation of the desired 
morphological sequence before using the updated
decoder state to attend over the character sequence
of the lemma. Secondly, in order to reduce the bias
of the decoder's language model, they hallucinate
two types of data that encourage common affixes and
character copying.  
Simply allowing the model to learn to copy 
characters for several epochs significantly 
out-performs the task baseline, while further
improvements are obtained through fine-tuning.
Making use of an adversarial language discriminator,
cross lingual gains are highly-correlated to
linguistic similarity, while augmenting the data
with hallucinated forms and multiple related target 
language further improves the model.

The system from IT-IST also attends separately
to tags and lemmas, using a gating mechanism
to interpolate the importance of the individual
attentions.  By combining the gated dual-head
attention with a SparseMax activation function,
they are able to jointly learn stem and affix 
modifications, improving significantly over the
baseline system.

The relative system performance is described in \autoref{tab:sub2team}, which shows the average per-language accuracy of each system. The table reflects the fact that some
teams submitted more than one system (e.g.~Tuebingen-1 \& Tuebingen-2 in the table).

\subsection{Task 2 Results}

\begin{table*}
    \centering
    \begin{adjustbox}{max width=\textwidth}
    \begin{tabular}{l r r r r}
\toprule
Team & Lemma Accuracy & Lemma Levenshtein & Morph Accuracy & Morph F1  \\
\midrule

CBNU-01$\dagger$ & 94.07 & 0.13 & 88.09 & 91.84\\
CHARLES-MALTA-01 & 74.95 & 0.62 & 50.37 & 58.81 \\
CHARLES-SAARLAND-02$\dagger$ & 95.00 & 0.11 & \textbf{93.23} & \textbf{96.02} \\
CMU-02 & 92.20 & 0.17 & 85.06 & 88.97 \\
CMU-DataAug-01$\ddagger$ & 92.51 & 0.17 & 86.53 & 91.18 \\
Edinburgh-01 & 94.20 & 0.13 & 88.93 & 92.89 \\
ITU-01 & 94.46 & 0.11 & 86.67 & 90.54 \\
NLPCUBE-01 & 91.43 & 2.43 & 84.92 & 88.67 \\
OHIOSTATE-01 & 93.43 & 0.17 & 87.42 & 92.51 \\
RUG-01$\dagger$ & 93.91 & 0.14 & 90.53 & 94.54 \\
RUG-02 & 93.06 & 0.15 & 88.80 & 93.22 \\
UFALPRAGUE-01$\dagger$ & \textbf{95.78} & \textbf{0.10} & 93.19 & 95.92 \\
UNTHILTLING-02$\dagger$ & 83.14 & 0.55 & 15.69  & 51.87 \\
EDINBURGH-02* & 97.35 & 0.06 & 93.02 & 95.94 \\
CMU-Monolingual* & 88.31 & 0.27 & 84.60  & 91.18 \\
CMU-PolyGlot-01*$\dagger$ & 76.81 & 0.54 & 60.98 & 75.42 \\ \hline
Baseline & 94.17 & 0.13 & 73.16 & 87.92 \\

\bottomrule

    \end{tabular}
    \end{adjustbox}
    \caption{Task 2 Team Scores, averaged across all treebanks; * indicates submissions were
    only applied to a subset of languages, making scores incomparable. $\dagger$ indicates that additional external resources were used for training, and $\ddagger$ indicates that training data were shared across languages or treebanks.}
    \label{tab:sub2team}
\end{table*}

\begin{table*}
    \centering
    \begin{adjustbox}{width=1.10\textwidth, angle=90}
    \begin{tabular}{l r r l | l r r l}
\toprule
Language (Treebank) & Baseline & Best & Team & Language (Treebank) & Baseline & Best & Team\\
\midrule
UD\_Afrikaans-AfriBooms & 98.41 & 99.15 & UFALPRAGUE-01  &                   UD\_Italian-PoSTWITA & 95.60 & 97.95 & UFALPRAGUE-01  \\
UD\_Akkadian-PISANDUB & 66.83 & 67.82 & CBNU-01 / EDINBURGH-01  &                   UD\_Italian-PUD & 95.59 & 98.06 & UFALPRAGUE-01  \\
UD\_Amharic-ATT & 98.68 & 100.00 & Multiple &                   UD\_Japanese-GSD & 97.71 & 99.65 & CHARLES-SAARLAND-02  \\
UD\_Ancient\_Greek-Perseus & 94.44 & 95.24 & EDINBURGH-01  &                   UD\_Japanese-Modern & 94.20& 98.67 & CHARLES-SAARLAND-02  \\
UD\_Ancient\_Greek-PROIEL & 96.68 & 97.49 & EDINBURGH-01  &                   UD\_Japanese-PUD & 95.75 & 99.36 & CHARLES-SAARLAND-02  \\
UD\_Arabic-PADT & 94.49 & 96.08 & UFALPRAGUE-01  &                   UD\_Komi\_Zyrian-IKDP & 78.91 & 89.84 & RUG-02  \\
UD\_Arabic-PUD & 85.24 & 87.13 & EDINBURGH-01  &                   UD\_Komi\_Zyrian-Lattice & 82.97 & 87.91 & UFALPRAGUE-01  \\
UD\_Armenian-ArmTDP & 95.39 & 95.96 & UFALPRAGUE-01  &                   UD\_Korean-GSD & 92.25 & 94.21 & UFALPRAGUE-01  \\
UD\_Bambara-CRB & 87.02 & 92.71 & UFALPRAGUE-01  &                   UD\_Korean-Kaist & 94.61 & 95.78 & EDINBURGH-01  \\
UD\_Basque-BDT & 96.07 & 97.19 & UFALPRAGUE-01  &                   UD\_Korean-PUD & 96.41 & 99.57 & CHARLES-SAARLAND-02  \\
UD\_Belarusian-HSE & 89.70 & 92.51 & CHARLES-SAARLAND-02  &                   UD\_Kurmanji-MG & 92.29 & 94.80 & UFALPRAGUE-01  \\
UD\_Breton-KEB & 93.53 & 93.83 & OHIOSTATE-01  &                   UD\_Latin-ITTB & 98.17 & 99.20 & CHARLES-SAARLAND-02  \\
UD\_Bulgarian-BTB & 97.37 & 98.36 & UFALPRAGUE-01  &                   UD\_Latin-Perseus & 89.54 & 93.49 & UFALPRAGUE-01  \\
UD\_Buryat-BDT & 88.56 & 90.19 & UFALPRAGUE-01  &                   UD\_Latin-PROIEL & 96.41 & 97.37 & UFALPRAGUE-01  \\
UD\_Cantonese-HK & 91.61 & 100.00 & Multiple &                   UD\_Latvian-LVTB & 95.59 & 97.23 & UFALPRAGUE-01  \\
UD\_Catalan-AnCora & 98.07 & 99.38 & CHARLES-SAARLAND-02  &                   UD\_Lithuanian-HSE & 86.42 & 87.44 & OHIOSTATE-01  \\
UD\_Chinese-CFL & 93.26 & 99.76 & CBNU-01 / UFALPRAGUE-01  &                   UD\_Marathi-UFAL & 75.61 & 76.69 & CHARLES-SAARLAND-02  \\
UD\_Chinese-GSD & 98.44 & 99.98 & CBNU-01 / CMU-02 / UFALPRAGUE-01  &                   UD\_Naija-NSC & 99.33 & 100.00 & Multiple\\
UD\_Coptic-Scriptorium & 95.80 & 97.31 & UFALPRAGUE-01  &                   UD\_North\_Sami-Giella & 93.04 & 93.47 & OHIOSTATE-01  \\
UD\_Croatian-SET & 95.32 & 97.52 & UFALPRAGUE-01  &                   UD\_Norwegian-Bokmaal & 98.00 & 99.19 & UFALPRAGUE-01  \\
UD\_Czech-CAC & 97.82 & 99.45 & CHARLES-SAARLAND-02  &                   UD\_Norwegian-Nynorsk & 97.85 & 99.00 & CHARLES-SAARLAND-02  \\
UD\_Czech-CLTT & 98.21 & 99.47 & UFALPRAGUE-01  &                   UD\_Norwegian-NynorskLIA & 96.66 & 98.22 & UFALPRAGUE-01  \\
UD\_Czech-FicTree & 97.66 & 99.01 & CHARLES-SAARLAND-02  &                   UD\_Old\_Church\_Slavonic-PROIEL & 96.38 & 97.23 & EDINBURGH-01  \\
UD\_Czech-PDT & 96.06 & 99.42 & CHARLES-SAARLAND-02  &                   UD\_Persian-Seraji & 96.08 & 96.89 & UFALPRAGUE-01  \\
UD\_Czech-PUD & 93.58 & 98.13 & UFALPRAGUE-01  &                   UD\_Polish-LFG & 95.82 & 97.94 & CHARLES-SAARLAND-02  \\
UD\_Danish-DDT & 96.16 & 98.33 & UFALPRAGUE-01  &                   UD\_Polish-SZ & 95.18 & 97.43 & CHARLES-SAARLAND-02  \\
UD\_Dutch-Alpino & 97.35 & 98.62 & CHARLES-SAARLAND-02  &                   UD\_Portuguese-Bosque & 97.08 & 98.69 & UFALPRAGUE-01  \\
UD\_Dutch-LassySmall & 96.63 & 98.21 & UFALPRAGUE-01  &                   UD\_Portuguese-GSD & 93.70 & 99.11 & UFALPRAGUE-01  \\
UD\_English-EWT & 97.68 & 99.19 & CHARLES-SAARLAND-02  &                   UD\_Romanian-Nonstandard & 95.86 & 96.74 & UFALPRAGUE-01  \\
UD\_English-GUM & 97.41 & 98.63 & UFALPRAGUE-01  &                   UD\_Romanian-RRT & 96.94 & 98.60 & UFALPRAGUE-01  \\
UD\_English-LinES & 98.00 & 98.62 & CHARLES-SAARLAND-02  &                   UD\_Russian-GSD & 95.67 & 97.77 & UFALPRAGUE-01  \\
UD\_English-ParTUT & 97.66 & 98.52 & UFALPRAGUE-01  &                   UD\_Russian-PUD & 91.85 & 95.76 & UFALPRAGUE-01  \\
UD\_English-PUD & 95.29 & 97.89 & CHARLES-SAARLAND-02  &                   UD\_Russian-SynTagRus & 95.92 & 99.01 & CHARLES-SAARLAND-02  \\
UD\_Estonian-EDT & 94.84 & 97.09 & EDINBURGH-01  &                   UD\_Russian-Taiga & 89.86 & 100.00 & UNTHILTLING-02  \\
UD\_Faroese-OFT & 88.86 & 89.53 & UFALPRAGUE-01  &                   UD\_Sanskrit-UFAL & 64.32 & 67.34 & CMU-Monolingual-01  \\
UD\_Finnish-FTB & 94.88 & 96.64 & EDINBURGH-02  &                   UD\_Serbian-SET & 96.72 & 98.19 & UFALPRAGUE-01  \\
UD\_Finnish-PUD & 88.27 & 89.98 & UFALPRAGUE-01  &                   UD\_Slovak-SNK & 96.14 & 97.57 & CHARLES-SAARLAND-02  \\
UD\_Finnish-TDT & 95.53 & 96.60 & UFALPRAGUE-01  &                   UD\_Slovenian-SSJ & 96.43 & 98.87 & CHARLES-SAARLAND-02  \\
UD\_French-GSD & 97.97 & 99.01 & CHARLES-SAARLAND-02  &                   UD\_Slovenian-SST & 94.06 & 97.20 & CHARLES-SAARLAND-02  \\
UD\_French-ParTUT & 95.69 & 96.66 & CHARLES-SAARLAND-02  &                   UD\_Spanish-AnCora & 98.54 & 99.46 & UFALPRAGUE-01  \\
UD\_French-Sequoia & 97.67 & 99.01 & UFALPRAGUE-01  &                   UD\_Spanish-GSD & 98.42 & 99.30 & UFALPRAGUE-01  \\
UD\_French-Spoken & 97.98 & 99.52 & post\_deadline\_RUG-01 &                   UD\_Swedish-LinES & 95.85 & 98.30 & UFALPRAGUE-01  \\
UD\_Galician-CTG & 98.22 & 98.96 & CHARLES-SAARLAND-02  &                   UD\_Swedish-PUD & 93.12 & 96.63 & UFALPRAGUE-01  \\
UD\_Galician-TreeGal & 96.18 & 98.65 & UFALPRAGUE-01  &                   UD\_Swedish-Talbanken & 97.23 & 98.62 & CHARLES-SAARLAND-02  \\
UD\_German-GSD & 96.26 & 97.65 & ITU-01  &                   UD\_Tagalog-TRG & 78.38 & 91.89 & Multiple \\
UD\_Gothic-PROIEL & 96.53 & 97.03 & EDINBURGH-01  &                   UD\_Tamil-TTB & 93.86 & 96.43 & UFALPRAGUE-01  \\
UD\_Greek-GDT & 96.76 & 97.24 & EDINBURGH-01  &                   UD\_Turkish-IMST & 96.41 & 96.84 & UFALPRAGUE-01  \\
UD\_Hebrew-HTB & 96.72 & 98.17 & UFALPRAGUE-01  &                   UD\_Turkish-PUD & 86.02 & 89.03 & UFALPRAGUE-01  \\
UD\_Hindi-HDTB & 98.60 & 98.87 & UFALPRAGUE-01  &                   UD\_Ukrainian-IU & 95.53 & 97.85 & UFALPRAGUE-01  \\
UD\_Hungarian-Szeged & 95.17 & 97.47 & UFALPRAGUE-01  &                   UD\_Upper\_Sorbian-UFAL & 91.69 & 93.74 & CHARLES-SAARLAND-02  \\
UD\_Indonesian-GSD & 99.37 & 99.61 & UFALPRAGUE-01  &                   UD\_Urdu-UDTB & 96.19 & 96.98 & UFALPRAGUE-01  \\
UD\_Irish-IDT & 91.69 & 92.02 & OHIOSTATE-01  &                   UD\_Vietnamese-VTB & 99.79 & 100.00 & CMU-02 / UNTHILTLING-02  \\
UD\_Italian-ISDT & 97.38 & 98.88 & CHARLES-SAARLAND-02 / UFALPRAGUE-01 &                   UD\_Yoruba-YTB & 98.84 & 98.84 & Multiple \\
UD\_Italian-ParTUT & 96.84 & 98.87 & CHARLES-SAARLAND-02  &                  \\
\bottomrule

    \end{tabular}
    \end{adjustbox}
    \caption{Task 2 Lemma Accuracy scores}
\end{table*}

\begin{table*}
    \centering
    \begin{adjustbox}{max width=1.20\linewidth, angle=90}
    \begin{tabular}{l r r l | l r r l}
\toprule
Language (Treebank) & Baseline & Best & Team & Language (Treebank) & Baseline & Best & Team \\
\midrule

UD\_Afrikaans-AfriBooms & 0.03 & 0.02 & Multiple &                             UD\_Italian-PoSTWITA & 0.11 & 0.05 & UFALPRAGUE-01 \\
UD\_Akkadian-PISANDUB & 0.87 & 0.85 & OHIOSTATE-01 &                           UD\_Italian-PUD & 0.08 & 0.04 & CHARLES-SAARLAND-02 / UFALPRAGUE-01  \\
UD\_Amharic-ATT & 0.02 & 0.00 & Multiple &                                     UD\_Japanese-GSD & 0.04 & 0.01 & Multiple \\                                
UD\_Ancient\_Greek-Perseus & 0.14 & 0.12 & EDINBURGH-01 &                      UD\_Japanese-Modern & 0.07 & 0.01 & CHARLES-SAARLAND-02  \\                 
UD\_Ancient\_Greek-PROIEL & 0.08 & 0.06 & EDINBURGH-01 / EDINBURGH-02 &        UD\_Japanese-PUD & 0.07 & 0.01 & CHARLES-SAARLAND-02 / UFALPRAGUE-01  \\
UD\_Arabic-PADT & 0.16 & 0.11 & UFALPRAGUE-01 &                                UD\_Komi\_Zyrian-IKDP & 0.38 & 0.23 & RUG-01 / RUG-02  \\                  
UD\_Arabic-PUD & 0.41 & 0.37 & EDINBURGH-01 &                                  UD\_Komi\_Zyrian-Lattice & 0.34 & 0.25 & UFALPRAGUE-01  \\               
UD\_Armenian-ArmTDP & 0.08 & 0.07 & UFALPRAGUE-01 &                            UD\_Korean-GSD & 0.18 & 0.11 & Multiple  \\                                
UD\_Bambara-CRB & 0.27 & 0.10 & UFALPRAGUE-01 &                                 UD\_Korean-Kaist & 0.09 & 0.06 & EDINBURGH-01  \\                           
UD\_Basque-BDT & 0.09 & 0.06 & UFALPRAGUE-01 &                                 UD\_Korean-PUD & 0.06 & 0.01 & Multiple  \\                               
UD\_Belarusian-HSE & 0.17 & 0.12 & CHARLES-SAARLAND-02 &                       UD\_Kurmanji-MG & 0.39 & 0.10 & UFALPRAGUE-01  \\                           
UD\_Breton-KEB & 0.16 & 0.13 & ITU-01 &                                     
UD\_Latin-ITTB & 0.04 & 0.02 & CHARLES-SAARLAND-02 / UFALPRAGUE-01  \\     
UD\_Bulgarian-BTB & 0.07 & 0.05 & ITU-01 / UFALPRAGUE-01 &                     UD\_Latin-Perseus & 0.21 & 0.13 & UFALPRAGUE-01  \\              
UD\_Buryat-BDT & 0.27 & 0.22 & UFALPRAGUE-01 &                                 UD\_Latin-PROIEL & 0.08 & 0.05 & CHARLES-SAARLAND-02  \\         
UD\_Cantonese-HK & 0.28 & 0.00 & Multiple &                                     UD\_Latvian-LVTB & 0.07 & 0.05 & CHARLES-SAARLAND-02 / UFALPRAGUE-01  \\
UD\_Catalan-AnCora & 0.04 & 0.01 & CHARLES-SAARLAND-02 / UFALPRAGUE-01 &       UD\_Lithuanian-HSE & 0.25 & 0.24 & UFALPRAGUE-01  \\                   
UD\_Chinese-CFL & 0.10 & 0.01 & NLPCUBE-01 &                                    UD\_Marathi-UFAL & 0.86 & 0.57 & CMU-Monolingual-01  \\
UD\_Chinese-GSD & 0.02 & 0.01 & Multiple &                                     
UD\_Naija-NSC & 0.01 & 0.00 & Multiple  \\
UD\_Coptic-Scriptorium & 0.09 & 0.06 & UFALPRAGUE-01 &                         UD\_North\_Sami-Giella & 0.14 & 0.13 & EDINBURGH-01 / OHIOSTATE-01  \\
UD\_Croatian-SET & 0.09 & 0.05 & CHARLES-SAARLAND-02 / UFALPRAGUE-01 &         UD\_Norwegian-Bokmaal & 0.03 & 0.01 & CHARLES-SAARLAND-02 / UFALPRAGUE-01 \\
UD\_Czech-CAC & 0.05 & 0.01 & CHARLES-SAARLAND-02 / UFALPRAGUE-01 &            UD\_Norwegian-Nynorsk & 0.04 & 0.01 & CHARLES-SAARLAND-02  \\
UD\_Czech-CLTT & 0.04 & 0.01 & CHARLES-SAARLAND-02 / UFALPRAGUE-01 &           UD\_Norwegian-NynorskLIA & 0.08 & 0.03 & UFALPRAGUE-01  \\
UD\_Czech-FicTree & 0.04 & 0.02 & CHARLES-SAARLAND-02 / UFALPRAGUE-01 &        UD\_Old\_Church\_Slavonic-PROIEL & 0.08 & 0.06 & EDINBURGH-01  \\
UD\_Czech-PDT & 0.06 & 0.01 & CHARLES-SAARLAND-02 / UFALPRAGUE-01 &            UD\_Persian-Seraji & 0.19 & 0.15 & UFALPRAGUE-01  \\
UD\_Czech-PUD & 0.10 & 0.03 & UFALPRAGUE-01 &                                   UD\_Polish-LFG & 0.08 & 0.04 & CHARLES-SAARLAND-02 / UFALPRAGUE-01  \\
UD\_Danish-DDT & 0.06 & 0.03 & CHARLES-SAARLAND-02 / UFALPRAGUE-01 &           
UD\_Polish-SZ & 0.08 & 0.04 & UFALPRAGUE-01  \\
UD\_Dutch-Alpino & 0.05 & 0.03 & CHARLES-SAARLAND-02 / UFALPRAGUE-01 &         UD\_Portuguese-Bosque & 0.05 & 0.02 & CHARLES-SAARLAND-02 / UFALPRAGUE-01  \\
UD\_Dutch-LassySmall & 0.06 & 0.03 & CHARLES-SAARLAND-02 / UFALPRAGUE-01 &     UD\_Portuguese-GSD & 0.18 & 0.05 & CHARLES-SAARLAND-02 / UFALPRAGUE-01  \\
UD\_English-EWT & 0.12 & 0.01 & CHARLES-SAARLAND-02 &                          UD\_Romanian-Nonstandard & 0.08 & 0.06 & Multiple  \\                 
UD\_English-GUM & 0.05 & 0.02 & CHARLES-SAARLAND-02 / UFALPRAGUE-01 &          UD\_Romanian-RRT & 0.05 & 0.02 & CHARLES-SAARLAND-02   \\
UD\_English-LinES & 0.04 & 0.02 & CHARLES-SAARLAND-02 / UFALPRAGUE-01 &        UD\_Russian-GSD & 0.07 & 0.04 & CHARLES-SAARLAND-02 / UFALPRAGUE-01  \\
UD\_English-ParTUT & 0.04 & 0.02 & CHARLES-SAARLAND-02 / UFALPRAGUE-01 &       UD\_Russian-PUD & 0.18 & 0.08 & CHARLES-SAARLAND-02 / UFALPRAGUE-01  \\
UD\_English-PUD & 0.07 & 0.03 & CHARLES-SAARLAND-02 &                          UD\_Russian-SynTagRus & 0.08 & 0.02 & CHARLES-SAARLAND-02 / UFALPRAGUE-01  \\
UD\_Estonian-EDT & 0.11 & 0.05 & EDINBURGH-01 &                                UD\_Russian-Taiga & 0.21 & 0.00 & UWTHILTLING  \\                     
UD\_Faroese-OFT & 0.20 & 0.18 & ITU-01 &                                     UD\_Sanskrit-UFAL & 0.85 & 0.82 & CMU-Monolingual-01  \\
UD\_Finnish-FTB & 0.11 & 0.08 & Multiple &                                     UD\_Serbian-SET & 0.06 & 0.03 & CHARLES-SAARLAND-02 / UFALPRAGUE-01  \\
UD\_Finnish-PUD & 0.24 & 0.18 & UFALPRAGUE-01 &                                UD\_Slovak-SNK & 0.06 & 0.04 & CHARLES-SAARLAND-02  \\
UD\_Finnish-TDT & 0.10 & 0.07 & UFALPRAGUE-01 &                                 UD\_Slovenian-SSJ & 0.06 & 0.02 & CHARLES-SAARLAND-02 / UFALPRAGUE-01  \\
UD\_French-GSD & 0.04 & 0.02 & Multiple &                                     UD\_Slovenian-SST & 0.12 & 0.05 & CHARLES-SAARLAND-02  \\
UD\_French-ParTUT & 0.07 & 0.05 & RUG-02 / post\_deadline\_RUG-01 &            UD\_Spanish-AnCora & 0.03 & 0.01 & Multiple  \\
UD\_French-Sequoia & 0.05 & 0.02 & CHARLES-SAARLAND-02 / UFALPRAGUE-01 &       UD\_Spanish-GSD & 0.03 & 0.01 & Multiple \\
UD\_French-Spoken & 0.04 & 0.01 & post\_deadline\_RUG-01 &                     UD\_Swedish-LinES & 0.08 & 0.03 & UFALPRAGUE-01  \\
UD\_Galician-CTG & 0.04 & 0.02 & Multiple &                                    UD\_Swedish-PUD & 0.10 & 0.05 & UFALPRAGUE-01  \\
UD\_Galician-TreeGal & 0.06 & 0.03 & CHARLES-SAARLAND-02 / UFALPRAGUE-01 &     UD\_Swedish-Talbanken & 0.05 & 0.02 & CHARLES-SAARLAND-02 / UFALPRAGUE-01  \\
UD\_German-GSD & 0.08 & 0.04 & ITU-01 &                                     
UD\_Tagalog-TRG & 0.49 & 0.19 & CHARLES-SAARLAND-02 / ITU-01  \\
UD\_Gothic-PROIEL & 0.07 & 0.06 & OHIOSTATE-01 &                               
UD\_Tamil-TTB & 0.14 & 0.07 & UFALPRAGUE-01  \\
UD\_Greek-GDT & 0.07 & 0.06 & EDINBURGH-01 &                                   UD\_Turkish-IMST & 0.08 & 0.06 & EDINBURGH-01 / ITU-01 / UFALPRAGUE-01  \\
UD\_Hebrew-HTB & 0.06 & 0.03 & UFALPRAGUE-01 &                                 UD\_Turkish-PUD & 0.34 & 0.28 & ITU-01  \\
UD\_Hindi-HDTB & 0.02 & 0.01 & Multiple &                                     UD\_Ukrainian-IU & 0.10 & 0.03 & CHARLES-SAARLAND-02  \\
UD\_Hungarian-Szeged & 0.10 & 0.05 & UFALPRAGUE-01 &                            UD\_Upper\_Sorbian-UFAL & 0.12 & 0.10 & CHARLES-SAARLAND-02  \\
UD\_Indonesian-GSD & 0.01 & 0.01 & Multiple &                                  
UD\_Urdu-UDTB & 0.07 & 0.06 & Multiple  \\
UD\_Irish-IDT & 0.18 & 0.16 & OHIOSTATE-01 &                                   UD\_Vietnamese-VTB & 0.02 & 0.00 & CMU-02 / UNTHILTLING  \\
UD\_Italian-ISDT & 0.05 & 0.02 & CHARLES-SAARLAND-02 / UFALPRAGUE-01 &         UD\_Yoruba-YTB & 0.01 & 0.01 & Multiple  \\               
UD\_Italian-ParTUT & 0.08 & 0.02 & CHARLES-SAARLAND-02 &    \\              
\bottomrule

    \end{tabular}
    \end{adjustbox}
    \caption{Task 2 Lemma Levenshtein scores}
\end{table*}

\begin{table*}
    \centering
    \begin{adjustbox}{angle=90, max width=1.0\linewidth}
    \begin{tabular}{l r r l | l r r l}
\toprule
Language (Treebank) & Baseline & Best & Team & Language (Treebank) & Baseline & Best & Team \\
\midrule

UD\_Afrikaans-AfriBooms & 84.90 & 99.23 & CHARLES-SAARLAND-02 / UFALPRAGUE-01 &        UD\_Italian-PoSTWITA & 70.09 & 96.88 & CHARLES-SAARLAND-02 \\
UD\_Akkadian-PISANDUB & 78.22 & 89.11 & CHARLES-SAARLAND-02 &        UD\_Italian-PUD & 80.78 & 96.37 & CHARLES-SAARLAND-02 \\
UD\_Amharic-ATT & 75.43 & 89.79 & UFALPRAGUE-01 &        UD\_Japanese-GSD & 85.47 & 98.41 & CHARLES-SAARLAND-02 \\
UD\_Ancient\_Greek-Perseus & 69.88 & 91.94 & UFALPRAGUE-01 &        UD\_Japanese-Modern & 94.94 & 97.47 & CHARLES-SAARLAND-02 \\
UD\_Ancient\_Greek-PROIEL & 84.55 & 92.94 & UFALPRAGUE-01 &        UD\_Japanese-PUD & 84.33 & 98.63 & UFALPRAGUE-01 \\
UD\_Arabic-PADT & 76.78 & 95.66 & CHARLES-SAARLAND-02 &        UD\_Komi\_Zyrian-IKDP & 35.94 & 75.78 & UFALPRAGUE-01 \\
UD\_Arabic-PUD & 63.07 & 85.04 & UFALPRAGUE-01 &        UD\_Komi\_Zyrian-Lattice & 45.05 & 69.78 & UFALPRAGUE-01 \\
UD\_Armenian-ArmTDP & 64.38 & 93.34 & UFALPRAGUE-01 &        UD\_Korean-GSD & 79.73 & 96.77 & CHARLES-SAARLAND-02 \\
UD\_Bambara-CRB & 76.99 & 93.93 & UFALPRAGUE-01 &        UD\_Korean-Kaist & 84.30 & 97.85 & CHARLES-SAARLAND-02 \\
UD\_Basque-BDT & 67.76 & 92.52 & UFALPRAGUE-01 &        UD\_Korean-PUD & 76.78 & 94.67 & CHARLES-SAARLAND-02 \\
UD\_Belarusian-HSE & 54.22 & 89.93 & CHARLES-SAARLAND-02 &        UD\_Kurmanji-MG & 68.10 & 85.57 & UFALPRAGUE-01 \\
UD\_Breton-KEB & 76.52 & 91.14 & UFALPRAGUE-01 &        UD\_Latin-ITTB & 77.68 & 97.64 & CHARLES-SAARLAND-02 \\
UD\_Bulgarian-BTB & 79.64 & 98.01 & CHARLES-SAARLAND-02 &        UD\_Latin-Perseus & 55.06 & 87.76 & UFALPRAGUE-01 \\
UD\_Buryat-BDT & 64.23 & 88.56 & UFALPRAGUE-01 &        UD\_Latin-PROIEL & 82.16 & 93.68 & CHARLES-SAARLAND-02 \\
UD\_Cantonese-HK & 68.57 & 94.29 & CHARLES-SAARLAND-02 &        UD\_Latvian-LVTB & 70.33 & 95.78 & CHARLES-SAARLAND-02 \\
UD\_Catalan-AnCora & 85.57 & 98.82 & CHARLES-SAARLAND-02 &        UD\_Lithuanian-HSE & 41.43 & 80.14 & UFALPRAGUE-01 \\
UD\_Chinese-CFL & 76.71 & 94.09 & UFALPRAGUE-01 &        UD\_Marathi-UFAL & 40.11 & 67.75 & CHARLES-SAARLAND-02 \\
UD\_Chinese-GSD & 75.97 & 97.13 & CHARLES-SAARLAND-02 &        UD\_Naija-NSC & 66.42 & 96.57 & UFALPRAGUE-01 \\
UD\_Coptic-Scriptorium & 87.73 & 96.22 & UFALPRAGUE-01 &        UD\_North\_Sami-Giella & 66.87 & 92.46 & CHARLES-SAARLAND-02 \\
UD\_Croatian-SET & 71.42 & 94.42 & UFALPRAGUE-01 &        UD\_Norwegian-Bokmaal & 81.27 & 98.25 & CHARLES-SAARLAND-02 \\
UD\_Czech-CAC & 77.26 & 98.48 & CHARLES-SAARLAND-02 &        UD\_Norwegian-Nynorsk & 81.75 & 98.11 & CHARLES-SAARLAND-02 \\
UD\_Czech-CLTT & 72.60 & 95.81 & UFALPRAGUE-01 &        UD\_Norwegian-NynorskLIA & 74.20 & 96.80 & CHARLES-SAARLAND-02 \\
UD\_Czech-FicTree & 68.34 & 97.13 & CHARLES-SAARLAND-02 &        UD\_Old\_Church\_Slavonic-PROIEL & 84.13 & 93.01 & UFALPRAGUE-01 \\
UD\_Czech-PDT & 76.70 & 98.54 & CHARLES-SAARLAND-02 &        UD\_Persian-Seraji & 86.84 & 98.31 & CHARLES-SAARLAND-02 / UFALPRAGUE-01 \\
UD\_Czech-PUD & 60.67 & 95.03 & UFALPRAGUE-01 &        UD\_Polish-LFG & 65.72 & 97.13 & CHARLES-SAARLAND-02 \\
UD\_Danish-DDT & 77.22 & 97.98 & CHARLES-SAARLAND-02 &        UD\_Polish-SZ & 63.15 & 95.11 & CHARLES-SAARLAND-02 \\
UD\_Dutch-Alpino & 82.07 & 98.12 & CHARLES-SAARLAND-02 &        UD\_Portuguese-Bosque & 78.05 & 96.22 & CHARLES-SAARLAND-02 \\
UD\_Dutch-LassySmall & 76.78 & 98.50 & CHARLES-SAARLAND-02 &        UD\_Portuguese-GSD & 83.87 & 99.03 & CHARLES-SAARLAND-02 \\
UD\_English-EWT & 80.17 & 97.85 & CHARLES-SAARLAND-02 &        UD\_Romanian-Nonstandard & 74.71 & 95.01 & CHARLES-SAARLAND-02 \\
UD\_English-GUM & 79.57 & 97.52 & CHARLES-SAARLAND-02 &        UD\_Romanian-RRT & 81.62 & 98.19 & CHARLES-SAARLAND-02 \\
UD\_English-LinES & 80.30 & 97.77 & CHARLES-SAARLAND-02 &        UD\_Russian-GSD & 63.37 & 94.92 & CHARLES-SAARLAND-02 \\
UD\_English-ParTUT & 80.31 & 96.65 & CHARLES-SAARLAND-02 &        UD\_Russian-PUD & 60.68 & 91.15 & CHARLES-SAARLAND-02 \\
UD\_English-PUD & 77.59 & 96.67 & CHARLES-SAARLAND-02 &        UD\_Russian-SynTagRus & 73.64 & 98.38 & CHARLES-SAARLAND-02 \\
UD\_Estonian-EDT & 74.03 & 97.23 & CHARLES-SAARLAND-02 &        UD\_Russian-Taiga & 52.06 & 92.09 & UFALPRAGUE-01 \\
UD\_Faroese-OFT & 65.32 & 87.70 & UFALPRAGUE-01 &        UD\_Sanskrit-UFAL & 29.65 & 50.75 & UFALPRAGUE-01 \\
UD\_Finnish-FTB & 72.89 & 96.85 & CHARLES-SAARLAND-02 &        UD\_Serbian-SET & 77.05 & 97.02 & CHARLES-SAARLAND-02 \\
UD\_Finnish-PUD & 70.07 & 95.62 & CHARLES-SAARLAND-02 / UFALPRAGUE-01 &        UD\_Slovak-SNK & 64.04 & 95.41 & CHARLES-SAARLAND-02 \\
UD\_Finnish-TDT & 74.84 & 97.15 & UFALPRAGUE-01 &        UD\_Slovenian-SSJ & 73.82 & 97.04 & UFALPRAGUE-01 \\
UD\_French-GSD & 84.20 & 98.31 & CHARLES-SAARLAND-02 &        UD\_Slovenian-SST & 69.57 & 92.76 & CHARLES-SAARLAND-02 \\
UD\_French-ParTUT & 81.67 & 95.78 & UFALPRAGUE-01 &        UD\_Spanish-AnCora & 84.35 & 98.79 & CHARLES-SAARLAND-02 \\
UD\_French-Sequoia & 81.50 & 98.15 & UFALPRAGUE-01 &        UD\_Spanish-GSD & 81.90 & 95.88 & CHARLES-SAARLAND-02 \\
UD\_French-Spoken & 94.48 & 98.60 & CHARLES-SAARLAND-02 &        UD\_Swedish-LinES & 76.93 & 94.75 & CHARLES-SAARLAND-02 \\
UD\_Galician-CTG & 86.65 & 98.44 & CHARLES-SAARLAND-02 &        UD\_Swedish-PUD & 79.97 & 95.85 & UFALPRAGUE-01 \\
UD\_Galician-TreeGal & 76.40 & 96.21 & CHARLES-SAARLAND-02 &        UD\_Swedish-Talbanken & 81.37 & 98.09 & CHARLES-SAARLAND-02 \\
UD\_German-GSD & 68.35 & 90.43 & CHARLES-SAARLAND-02 &        UD\_Tagalog-TRG & 67.57 & 91.89 & CHARLES-SAARLAND-02 / UFALPRAGUE-01 \\
UD\_Gothic-PROIEL & 81.00 & 91.02 & CHARLES-SAARLAND-02 &        UD\_Tamil-TTB & 73.33 & 91.63 & UFALPRAGUE-01 \\
UD\_Greek-GDT & 77.44 & 95.95 & UFALPRAGUE-01 &        UD\_Turkish-IMST & 62.94 & 92.27 & UFALPRAGUE-01 \\
UD\_Hebrew-HTB & 81.15 & 97.67 & CHARLES-SAARLAND-02 &        UD\_Turkish-PUD & 66.30 & 87.63 & post\_deadline\_RUG-01 \\
UD\_Hindi-HDTB & 80.60 & 93.65 & CHARLES-SAARLAND-02 &        UD\_Ukrainian-IU & 63.59 & 95.78 & CHARLES-SAARLAND-02 \\
UD\_Hungarian-Szeged & 65.90 & 95.03 & UFALPRAGUE-01 &        UD\_Upper\_Sorbian-UFAL & 57.70 & 87.02 & UFALPRAGUE-01 \\
UD\_Indonesian-GSD & 71.73 & 92.48 & CHARLES-SAARLAND-02 &        UD\_Urdu-UDTB & 69.97 & 80.90 & UFALPRAGUE-01 \\
UD\_Irish-IDT & 67.66 & 86.37 & UFALPRAGUE-01 &        UD\_Vietnamese-VTB & 69.42 & 94.54 & CHARLES-SAARLAND-02 \\
UD\_Italian-ISDT & 83.72 & 98.49 & CHARLES-SAARLAND-02 &        UD\_Yoruba-YTB & 73.26 & 93.80 & CMU-DataAug-01 \\
UD\_Italian-ParTUT & 83.51 & 98.72 & UFALPRAGUE-01 &       \\
\bottomrule

    \end{tabular}
    \end{adjustbox}
    \caption{Task 2 Morph Accuracy scores}
\end{table*}

\begin{table*}
    \centering
    \begin{adjustbox}{max width=1.1\linewidth, angle=90}
    \begin{tabular}{l r r l | l r r l}
\toprule
Language (Treebank) & Baseline & Best & Team & Language (Treebank) & Baseline & Best & Team \\
\midrule

UD\_Afrikaans-AfriBooms & 92.87 & 99.40 & UFALPRAGUE-01 &                    UD\_Italian-PoSTWITA & 87.98 & 97.90 & CHARLES-SAARLAND-02 \\
UD\_Akkadian-PISANDUB & 80.41 & 89.06 & CHARLES-SAARLAND-02 &                    UD\_Italian-PUD & 92.24 & 98.42 & CHARLES-SAARLAND-02 \\
UD\_Amharic-ATT & 87.57 & 93.15 & UFALPRAGUE-01 &                    UD\_Japanese-GSD & 90.64 & 98.21 & CHARLES-SAARLAND-02 \\
UD\_Ancient\_Greek-Perseus & 88.97 & 96.72 & UFALPRAGUE-01 &                    UD\_Japanese-Modern & 95.64 & 97.50 & CHARLES-SAARLAND-02 \\
UD\_Ancient\_Greek-PROIEL & 93.55 & 97.88 & UFALPRAGUE-01 &                    UD\_Japanese-PUD & 89.64 & 98.49 & UFALPRAGUE-01 \\
UD\_Arabic-PADT & 91.82 & 97.65 & CHARLES-SAARLAND-02 &                    UD\_Komi\_Zyrian-IKDP & 59.52 & 82.99 & UFALPRAGUE-01 \\
UD\_Arabic-PUD & 86.35 & 94.66 & RUG-01 &                    UD\_Komi\_Zyrian-Lattice & 74.12 & 82.99 & RUG-01 / RUG-02 \\
UD\_Armenian-ArmTDP & 86.74 & 96.66 & CHARLES-SAARLAND-02 &                    UD\_Korean-GSD & 85.90 & 96.27 & CHARLES-SAARLAND-02 \\
UD\_Bambara-CRB & 88.94 & 95.55 & UFALPRAGUE-01 &                    UD\_Korean-Kaist & 89.45 & 97.58 & CHARLES-SAARLAND-02 \\
UD\_Basque-BDT & 87.54 & 96.30 & CHARLES-SAARLAND-02 &                    UD\_Korean-PUD & 88.15 & 96.76 & CHARLES-SAARLAND-02 \\
UD\_Belarusian-HSE & 78.80 & 95.68 & CHARLES-SAARLAND-02 &                    UD\_Kurmanji-MG & 86.54 & 91.28 & UFALPRAGUE-01 \\
UD\_Breton-KEB & 88.34 & 93.79 & UFALPRAGUE-01 &                    UD\_Latin-ITTB & 93.12 & 98.96 & CHARLES-SAARLAND-02 \\
UD\_Bulgarian-BTB & 93.85 & 99.18 & CHARLES-SAARLAND-02 &                    UD\_Latin-Perseus & 78.91 & 94.65 & UFALPRAGUE-01 \\
UD\_Buryat-BDT & 80.94 & 90.50 & UFALPRAGUE-01 &                    UD\_Latin-PROIEL & 91.42 & 97.87 & CHARLES-SAARLAND-02 \\
UD\_Cantonese-HK & 76.80 & 92.83 & CHARLES-SAARLAND-02 &                    UD\_Latvian-LVTB & 89.55 & 98.04 & CHARLES-SAARLAND-02 \\
UD\_Catalan-AnCora & 95.73 & 99.45 & CHARLES-SAARLAND-02 &                    UD\_Lithuanian-HSE & 67.39 & 87.97 & CHARLES-SAARLAND-02 \\
UD\_Chinese-CFL & 82.05 & 93.21 & UFALPRAGUE-01 &                    UD\_Marathi-UFAL & 69.71 & 80.19 & CHARLES-SAARLAND-02 \\
UD\_Chinese-GSD & 83.79 & 97.04 & CHARLES-SAARLAND-02 &                    UD\_Naija-NSC & 76.73 & 95.47 & UFALPRAGUE-01 \\
UD\_Coptic-Scriptorium & 93.56 & 97.17 & UFALPRAGUE-01 &                    UD\_North\_Sami-Giella & 85.45 & 95.33 & CHARLES-SAARLAND-02 \\
UD\_Croatian-SET & 90.39 & 97.82 & CHARLES-SAARLAND-02 &                    UD\_Norwegian-Bokmaal & 93.17 & 99.02 & CHARLES-SAARLAND-02 \\
UD\_Czech-CAC & 93.94 & 99.48 & CHARLES-SAARLAND-02 &                    UD\_Norwegian-Nynorsk & 92.85 & 98.97 & CHARLES-SAARLAND-02 \\
UD\_Czech-CLTT & 92.61 & 98.32 & UFALPRAGUE-01 &                    UD\_Norwegian-NynorskLIA & 89.21 & 97.39 & CHARLES-SAARLAND-02 \\
UD\_Czech-FicTree & 90.32 & 98.90 & CHARLES-SAARLAND-02 &                    UD\_Old\_Church\_Slavonic-PROIEL & 91.17 & 97.13 & UFALPRAGUE-01 \\
UD\_Czech-PDT & 94.23 & 99.47 & CHARLES-SAARLAND-02 &                    UD\_Persian-Seraji & 93.76 & 98.68 & UFALPRAGUE-01 \\
UD\_Czech-PUD & 85.73 & 98.23 & UFALPRAGUE-01 &                    UD\_Polish-LFG & 88.73 & 98.86 & CHARLES-SAARLAND-02 \\
UD\_Danish-DDT & 90.19 & 98.68 & CHARLES-SAARLAND-02 &                    UD\_Polish-SZ & 86.24 & 98.11 & CHARLES-SAARLAND-02 \\
UD\_Dutch-Alpino & 91.25 & 98.62 & CHARLES-SAARLAND-02 &                    UD\_Portuguese-Bosque & 92.36 & 98.26 & CHARLES-SAARLAND-02 \\
UD\_Dutch-LassySmall & 87.97 & 98.83 & CHARLES-SAARLAND-02 &                    UD\_Portuguese-GSD & 91.73 & 99.10 & CHARLES-SAARLAND-02 \\
UD\_English-EWT & 90.91 & 98.52 & CHARLES-SAARLAND-02 &                    UD\_Romanian-Nonstandard & 91.70 & 97.65 & CHARLES-SAARLAND-02 \\
UD\_English-GUM & 89.81 & 98.11 & CHARLES-SAARLAND-02 &                    UD\_Romanian-RRT & 93.88 & 98.89 & CHARLES-SAARLAND-02 \\
UD\_English-LinES & 90.58 & 98.30 & CHARLES-SAARLAND-02 &                    UD\_Russian-GSD & 87.49 & 97.95 & CHARLES-SAARLAND-02 \\
UD\_English-ParTUT & 89.46 & 97.35 & CHARLES-SAARLAND-02 &                    UD\_Russian-PUD & 84.31 & 96.27 & CHARLES-SAARLAND-02 \\
UD\_English-PUD & 87.70 & 97.58 & CHARLES-SAARLAND-02 &                    UD\_Russian-SynTagRus & 92.73 & 99.23 & CHARLES-SAARLAND-02 \\
UD\_Estonian-EDT & 91.52 & 98.69 & CHARLES-SAARLAND-02 &                    UD\_Russian-Taiga & 76.77 & 95.56 & UFALPRAGUE-01 \\
UD\_Faroese-OFT & 85.73 & 93.98 & UFALPRAGUE-01 &                    UD\_Sanskrit-UFAL & 57.80 & 69.63 & RUG-01 / RUG-02 \\
UD\_Finnish-FTB & 89.08 & 98.38 & CHARLES-SAARLAND-02 &                    UD\_Serbian-SET & 91.75 & 98.64 & CHARLES-SAARLAND-02 \\
UD\_Finnish-PUD & 87.77 & 97.98 & CHARLES-SAARLAND-02 &                    UD\_Slovak-SNK & 88.04 & 98.24 & CHARLES-SAARLAND-02 \\
UD\_Finnish-TDT & 90.66 & 98.54 & CHARLES-SAARLAND-02 &                    UD\_Slovenian-SSJ & 90.12 & 98.80 & CHARLES-SAARLAND-02 \\
UD\_French-GSD & 94.63 & 99.07 & CHARLES-SAARLAND-02 &                    UD\_Slovenian-SST & 82.28 & 96.20 & CHARLES-SAARLAND-02 \\
UD\_French-ParTUT & 92.19 & 97.97 & UFALPRAGUE-01 &                    UD\_Spanish-AnCora & 95.35 & 99.40 & CHARLES-SAARLAND-02 \\
UD\_French-Sequoia & 93.04 & 99.11 & UFALPRAGUE-01 &                    UD\_Spanish-GSD & 93.95 & 98.08 & CHARLES-SAARLAND-02 \\
UD\_French-Spoken & 94.80 & 98.65 & CHARLES-SAARLAND-02 &                    UD\_Swedish-LinES & 89.99 & 97.67 & CHARLES-SAARLAND-02 \\
UD\_Galician-CTG & 91.35 & 98.29 & CHARLES-SAARLAND-02 &                    UD\_Swedish-PUD & 90.49 & 97.40 & UFALPRAGUE-01 \\
UD\_Galician-TreeGal & 89.33 & 97.88 & CHARLES-SAARLAND-02 &                    UD\_Swedish-Talbanken & 92.65 & 99.05 & CHARLES-SAARLAND-02 \\
UD\_German-GSD & 88.91 & 95.90 & CHARLES-SAARLAND-02 &                    UD\_Tagalog-TRG & 87.07 & 95.04 & CHARLES-SAARLAND-02 / UFALPRAGUE-01 \\
UD\_Gothic-PROIEL & 90.02 & 96.64 & CHARLES-SAARLAND-02 &                    UD\_Tamil-TTB & 89.22 & 96.00 & UFALPRAGUE-01 \\
UD\_Greek-GDT & 93.45 & 98.37 & UFALPRAGUE-01 &                    UD\_Turkish-IMST & 86.10 & 96.30 & UFALPRAGUE-01 \\
UD\_Hebrew-HTB & 91.79 & 98.47 & CHARLES-SAARLAND-02 &                    UD\_Turkish-PUD & 87.62 & 94.96 & post\_deadline\_RUG-01 \\
UD\_Hindi-HDTB & 93.92 & 98.04 & CHARLES-SAARLAND-02 &                    UD\_Ukrainian-IU & 86.81 & 98.10 & CHARLES-SAARLAND-02 \\
UD\_Hungarian-Szeged & 87.62 & 98.25 & UFALPRAGUE-01 &                    UD\_Upper\_Sorbian-UFAL & 81.04 & 93.51 & UFALPRAGUE-01 \\
UD\_Indonesian-GSD & 86.12 & 95.16 & CHARLES-SAARLAND-02 &                    UD\_Urdu-UDTB & 89.46 & 93.45 & CHARLES-SAARLAND-02 \\
UD\_Irish-IDT & 81.58 & 91.46 & UFALPRAGUE-01 &                    UD\_Vietnamese-VTB & 78.00 & 94.02 & CHARLES-SAARLAND-02 \\
UD\_Italian-ISDT & 94.46 & 99.19 & CHARLES-SAARLAND-02 &                    UD\_Yoruba-YTB & 85.47 & 94.19 & CMU-DataAug-01 \\
UD\_Italian-ParTUT & 93.88 & 99.21 & UFALPRAGUE-01 &                    \\\bottomrule

    \end{tabular}
    \end{adjustbox}
    \caption{Task 2 Morph F1 scores}
\end{table*}

Nine teams submitted system papers for Task 2,
with several interesting modifications to either the baseline or other prior work that
led to modest improvements.

Charles-Saarland achieved the
highest overall tagging accuracy by leveraging
multi-lingual BERT embeddings fine-tuned on
a concatenation of all available languages,
effectively transporting the cross-lingual objective
of Task 1 into Task 2.  Lemmas and tags are 
decoded separately (with a joint encoder and 
separate attention); Lemmas 
are a sequence of edit-actions, while tags are
calculated jointly. (There is no splitting of tags into features; tags are atomic.)

CBNU instead lemmatize using a 
transformer network, while performing tagging with
a multilayer perceptron with biaffine attention.
Input words are first lemmatized, and then pipelined
to the tagger, which produces atomic tag sequences (i.e., no splitting of features).

The team from Istanbul Technical University (ITU) jointly produces lemmatic 
edit-actions and morphological tags via a two level 
encoder (first word embeddings, and then 
context embeddings) and separate 
decoders.  Their system slightly improves over
the baseline lemmatization, but significantly 
improves tagging accuracy.

The team from the University of Groningen (RUG) also uses separate decoders
for lemmatization and tagging, but uses ELMo
to initialize the contextual embeddings, leading
to large gains in performance.  Furthermore,
joint training on related languages further improves
results.

CMU approaches tagging 
differently than the multi-task decoding we've
seen so far (baseline is used for lemmatization).  
Making use of a hierarchical CRF that first predicts
POS (that is subsequently looped back into the 
encoder), they then seek to predict each feature 
separately.  In particular, predicting POS 
separately greatly improves results. An attempt to leverage gold typological information led to little gain in the results; experiments suggest that the system is already learning the pertinent information.

The team from Ohio State University (OHIOSTATE) concentrates on predicting tags; the baseline lemmatizer is used for lemmatization.  To that end, they make use of 
a dual decoder that first predicts features
given only the word embedding as input; 
the predictions are
fed to a GRU seq2seq, which then predicts the
sequence of tags. 

The UNT HiLT+Ling team investigates a low-resource
setting of the tagging, by using parallel Bible data
to learn a translation matrix between English
and the target language, learning morphological
tags through analogy with English.

The UFAL-Prague team extends their submission from
the UD shared task (multi-layer LSTM), 
replacing the pretrained 
embeddings with BERT, to great success (first in 
lemmatization, 2nd in tagging).  Although they
predict complete tags, they use the individual 
features to regularize the decoder.  Small gains are also obtained
from joining multi-lingual corpora and ensembling.

CUNI--Malta performs lemmatization
as operations over edit actions with LSTM and ReLU.
Tagging is a bidirectional LSTM augmented by 
the edit actions (i.e., two-stage decoding), 
predicting features separately.

The Edinburgh system is a character-based LSTM encoder-decoder with
attention, implemented in OpenNMT. It can be seen as an extension of
the contextual lemmatization system Lematus \cite{bergmanis-goldwater-2018-context} 
to include morphological tagging, or alternatively as an
adaptation of the morphological re-inflection system MED \cite{kann-schutze-2016-med} to incorporate context and perform analysis rather than
re-inflection. Like these systems it uses a completely generic
encoder-decoder architecture with no specific adaptation to the
morphological processing task other than the form of the input. In the
submitted version of the system, the input is split into short chunks
corresponding to the target word plus one word of context on either
side, and the system is trained to output the corresponding lemmas and
tags for each three-word chunk.

Several teams relied on external resources to improve their lemmatization and feature analysis. Several teams made use of pre-trained embeddings. CHARLES-SAARLAND-2 and UFALPRAGUE-1 used pretrained contextual embeddings (BERT) provided by Google \citep{bert}. CBNU-1 used a mix of pre-trained embeddings from the CoNLL 2017 shared task and fastText.
Further, some teams trained their own embeddings to aid performance.

\section{Future Directions}

In general, the application of typology to natural language processing \citep[e.g.,][]{gerz-etal-2018-language,ponti2018modeling} provides an interesting avenue for multilinguality. Further, our shared task was designed to only leverage a single helper language, though many may exist with lexical or morphological overlap with the target language. Techniques like those of \citet{neubig-hu-2018-rapid} may aid in designing universal inflection architectures. Neither task this year included unannotated monolingual corpora. Using such data is well-motivated from an L1-learning point of view, and may affect the performance of low-resource data settings.

In the case of inflection an interesting future topic could involve departing from orthographic representation and using more IPA-like representations, i.e.\ transductions over pronunciations.  Different languages, in particular those with idiosyncratic orthographies, may offer new challenges in this respect.\footnote{Although some work suggests that working with IPA or phonological distinctive features in this context yields very similar results to working with graphemes \cite{wiemerslage-etal-2018-phonological}.}

Only one team tried to learn inflection in a multilingual setting---i.e.\ to use all training data to train one model. Such transfer learning is an interesting avenue of future research, but evaluation could be difficult. Whether any cross-language transfer is actually being learned vs.\ whether having more data better biases the networks to copy strings is an evaluation step to disentangle.\footnote{This has been addressed by \newcite{jin-kann-2017-exploring}.} 

Creating new data sets that accurately reflect learner exposure (whether L1 or L2) is also an important consideration in the design of future shared tasks. One pertinent facet of this is information about inflectional categories---often the inflectional information is insufficiently prescribed by the lemma, as with the Romanian verbal inflection classes or nominal gender in German.

As we move toward multilingual models for morphology, it becomes important to understand which representations are critical or irrelevant for adapting to new languages; this may be probed in the style of \citep{thompson-etal-2018-freezing}, and it can be used as a first step toward designing systems that avoid \say{catastrophic forgetting} as they learn to inflect new languages \citep{thompson-etal-2019-overcoming}.

Future directions for Task 2 include exploring cross-lingual analysis---in stride with both Task 1 and \citet{malaviya-etal-2018-neural}---and leveraging these analyses in downstream tasks.

\section{Conclusions}

The SIGMORPHON 2019 shared task provided a type-level evaluation on 100 language pairs in 79 languages and a token-level evaluation on 107 treebanks in 66 languages, of systems for inflection and analysis. On task 1 (low-resource inflection with cross-lingual transfer), 14 systems were submitted, while on task 2 (lemmatization and morphological feature analysis), 16 systems were submitted. All used neural network models, completing a trend in past years' shared tasks and other recent work on morphology.

%\todo[inline]{Write about actual findings from the task.}

In task 1, gains from cross-lingual
training were generally modest, with
gains positively correlating with
the linguistic similarity of the two
languages.

In the second task, several methods were
implemented by multiple groups, with
the most successful systems implementing
variations of multi-headed attention,
multi-level encoding, multiple decoders,
and ELMo and BERT contextual embeddings.

We have released the training, development, and test sets, and expect these datasets to provide a useful benchmark for future research into learning of inflectional morphology and string-to-string transduction.

\section*{Acknowledgments}

MS has received funding from the European Research Council (ERC) under the European Union's Horizon 2020 research and innovation programme (grant agreement No 771113).

\bibliography{naaclhlt2019}
\bibliographystyle{acl_natbib}

\end{document}